\documentclass[nai]{iosart2x}
\usepackage{todonotes}
\usepackage{booktabs}
\usepackage{url}

\pubyear{0000}
\volume{0}
\firstpage{1}
\lastpage{1}

\usepackage{xspace}
\newcommand{\ie}{\textit{i}.\textit{e}.,\xspace}
\newcommand{\eg}{\textit{e}.\textit{g}.,\xspace}

\newcommand{\etal}{\textit{\etal}\xspace}

\begin{document}

\begin{frontmatter}

\title{What can knowledge graph alignment gain with Neuro-Symbolic learning approaches? }
\runtitle{What can knowledge graph alignment gain with Neuro-Symbolic approaches?}


\begin{aug}
\author[A]{\inits{P.G.}\fnms{Pedro Giesteira} \snm{Cotovio}\ead[label=e1]{pgcotovio@fc.ul.pt}%
\thanks{Corresponding author. \printead{e1}.}}
\author[B,C]{\inits{E.}\fnms{Ernesto} \snm{Jimenez-Ruiz}\ead[label=e3]{ernesto.jimenez-ruiz@city.ac.uk}}
\author[A]{\inits{C.}\fnms{Catia} \snm{Pesquita}\ead[label=e2]{clpesquita@fc.ul.pt}}
\address[A]{LASIGE, \orgname{Faculdade de Ciências da Universidade de Lisboa},
Lisboa, \cny{Portugal}\printead[presep={\\}]{e1,e2}}
\address[B]{Department of Computer Science
City,  \orgname{University of London},
London, \cny{UK}\printead[presep={\\}]{e3}}
\address[C]{SIRIUS, \orgname{University of Oslo},
Oslo, \cny{Norway}}
\end{aug}


\begin{abstract}
Knowledge Graphs (KG) are the backbone of many data-intensive applications since they can represent data coupled with its meaning and context. Aligning KGs across different domains and providers is necessary to afford a fuller and integrated representation.  
A severe limitation of current KG alignment (KGA) algorithms is that they fail to articulate logical thinking and reasoning with lexical, structural, and semantic data learning. Deep learning models are increasingly popular for KGA inspired by their good performance in other tasks, but they suffer from limitations in explainability, reasoning, and data efficiency.
Hybrid neurosymbolic learning models hold the promise of integrating logical and data perspectives to produce high-quality alignments that are explainable and support validation through human-centric approaches. This paper examines the current state of the art in KGA and explores the potential for neurosymbolic integration, highlighting promising research directions for combining these fields.

\end{abstract}

\begin{keyword}
\kwd{Knowledge Graph Alignment}
\kwd{Neuro-Symbolic models}
\kwd{Explainable Artificial Intelligence}
\end{keyword}

\end{frontmatter}


\section{Introduction}\label{intro}


With the increasing amount of information available, the need for an effective way to integrate and manage this information has become critical. Knowledge graphs (KGs) have emerged as a solution to this problem, providing graph-based data models that formally represent data objects typically using ontologies. These models can capture knowledge in a graph-based abstraction that presents a significant advantage from a practical point of view compared to other data models \cite{gutierrez2021}. Deductive and inductive reasoning can increase the knowledge base through simple logical rules or combined with machine learning methods like embeddings \cite{kolyvakis2020}. KGs are, therefore, an effective way of managing, integrating, and extracting knowledge and value from different data sources.

However, in most practical cases, a single KG is often incomplete, and its usefulness, especially in multi-domain settings, depends on the integration with other KGs. The performance of state-of-the-art KG alignment (KGA) algorithms is often only achieved through hand-crafted configuration since heterogeneous knowledge requires context that lexical, structural, and rule-based methods cannot always grasp \cite{gutierrez2021}. The goal of KGA is to automatically identify and merge semantically equivalent entities, relationships, and attributes across multiple KGs \cite{euzenat2007}. This process requires comparing the entities and relationships in one graph with those in another and determining their similarity. The resulting correspondences can be used to create a unified representation of the knowledge.


Recent subsymbolic approaches, including linguistic and structural models like Attention Networks and Graph Neural Networks, have been employed to address knowledge graph alignment \cite{he2022, chen2021, Hao2021, Qu2019, wang2018}. Although these approaches have shown promise, they lack certain advantages of logical models, such as explainability and robust deduction~\cite{besold2017}. In the Artificial Intelligence (AI) community, there is a growing recognition that augmenting subsymbolic AI with symbolic properties can effectively address its current limitations \cite{garcez2023, hitzler2022, kautz2022}. This convergence of symbolic and subsymbolic approaches, known as Neurosymbolic AI, holds significant potential for enhancing the capabilities of AI systems. By combining symbolic reasoning, which excels at knowledge representation and logical inference, with the power of subsymbolic methods in handling large-scale data and extracting patterns, Neurosymbolic AI could address many of the current challenges in KGA.
Hybrid networks like Logic Tensor Networks (LTNs) \cite{badreddine2022}, which tightly integrate neural learning with logical reasoning, show particular promise in addressing the limitations of current deep learning algorithms.

In this paper, we provide a more detailed overview of the current state-of-the-art in KGA and neurosymbolic integration and explore the challenges and research opportunities that neurosymbolic models offer to the field~of~KGA.

\section{Knowledge Graph Alignment}    

\subsection{Problem Definition}\label{problem_def}

A Knowledge Graph (KG) is a multi-relational graph of structured knowledge resources, commonly expressed by RDF (Resource Description Framework) triples of the form  <subject, predicate, object>, that can be formally structured according to an $OWL$ ontology\footnote{When referring to OWL we mean $OWL~2$.} \cite{chen2021}.


An $OWL$ ontology, represented as $O$, is a structured representation of entities within a specific domain. In addition, an ontology $O$ includes a set of valid names for the entities present, known as its signature, $Sig(O)$. Formally, an $OWL$ ontology can be seen as a collection of axioms that follow the syntactic rules and constraints defined by their underlying description logics \cite{Horrocks2006}. 

The signature, $Sig(O)$, is structured as the disjoint union of four finite entity sets: (i) $N_C$, a collection of unary symbols referred to as named concepts, (ii) $N_R$, a group of binary symbols known as named object properties, (iii) $N_D$, an assemblage of binary symbols designated as data properties, and (iv) $N_I$, a compilation of constant symbols identified as named individuals. 



When aligning two KGs, the objective is to identify a mapping function, denoted as $f$, that maps entities from one KG's signature to entities in another while preserving the structural properties of both KGs. 
Consider two KGs, $KG_1$, $KG_2$,  along with their respective signatures,  $Sig(KG_1)$ and $Sig(KG_2)$. A mapping between entities of $KG_1$, $KG_2$ is typically represented as a tuple  $<e_1, e_2, r, c>$ such that $e_1 \in Sig(KG_1)$ and $e_2 \in Sig(KG_2)$, $r \in \{\sqsubseteq, \sqsupseteq, \equiv\}$ is a semantic relation, and $c$ is a confidence value (normally $c \in [0, 1]$). 

With this understanding, we can conceptualize $f$ as a mapping function denoted as $f: Sig(KG_1) \rightarrow Sig(KG_2)$ that associates the entities in $KG_1$ with the corresponding entities in $KG_2$, and produces a set of mappings $M$, also referred to as an alignment.
Finally, the knowledge graph $KG_M$, which results from the union of $KG_1$, $KG_2$, and $M$, is called the aligned knowledge graph.

\subsubsection{Consistency Principle Violations}

The aligned knowledge graph, represented as $KG_M$, has the potential to produce axioms not directly derivable from its sources, $KG$, $KG'$, or its alignment, $M$, in isolation. The deductive difference~\cite{Konev2008} between $KG$ and $KG'$, denoted $diff_{\Sigma}(KG, KG')$, is defined as the set of axioms with signature $\Sigma = Sig(KG) \cup Sig(KG')$ that do not hold over $KG$ but are valid over $KG'$. 

Evidence of violations in consistency may signal issues with the mappings or discrepancies between the original KGs. These violations invariably culminate in aligned KGs that are either inconsistent or incoherent, or in other words, KGs comprising unsatisfiable concepts.

The consistency principle~\cite{solimando2017} stipulates that every named concept within the aligned knowledge graph $KG_M$ should be satisfiable, under the assumption that the union of the input KGs, represented as $KG^\emptyset = KG \cup KG'$, is devoid of unsatisfiable concepts. Accordingly, if the condition $(A \sqsubseteq \bot) \in diff_{\Sigma}(KG^\emptyset, KG_M)$ holds for any $A \in \Sigma$, the alignment $M$ is deemed violate the consistency principle. Therefore, a mapping m $\in$ M is said to be involved in the cause of an unsatisfiability $A \sqsubseteq \bot$ if there exists a subset of axioms $\mathcal{K} \subseteq KG_M$ such that  $\mathcal{K} \not\models A \sqsubseteq \bot$ and $\mathcal{K}\cup \{m\} \models A \sqsubseteq \bot$.




To validate this principle, we can define a function, $consist: e' \rightarrow \{0, 1\}$, such that if any mapping $f: e'\rightarrow f(e')$ is involved in the cause of unsatisfiability, then the function $consist(e')$ will output 0; otherwise, it will yield 1.

This approach can be considered as a stringent or "hard" since a single instance of unsatisfiability results in an output of zero. In contrast, a "soft" alternative approach can be defined as $softconsist: e' \rightarrow [0, 1]$, wherein $\bot_{e'}$ denotes the number of instances of unsatisfiability involving  $e'$,  such that $\lim_{{\bot_{e'} \to \infty}} softconsist(e') = 0$. In this case, the output approaches zero, as the occurrences of unsatisfiability tend towards infinity. 

One potential heuristic to guide the behaviour of the $softconsist$ function could involve employing the logistic function of the occurrences of unsatisfiability in the following way:

$$softconsist(e') = \frac{1}{1 + \mathrm{e}^{-\bot_{e'}}} $$


\subsubsection{Unsupervised Setting}\label{unsup_learning}

The majority of existing KG alignment approaches are fully unsupervised and thus do not require any previously created alignment to learn mapping functions. In practice, within an unsupervised learning setting, the mapping process is typically broken down into two components: a candidate generation function $G$, which produces a range of possible mappings between entities, and a filtering function $F$ designed to eliminate improbable or irrelevant mappings from the generated candidates. 


The candidate generation function $G$ uses different types of information, such as labels, descriptions, types, and relationships with other objects, to pinpoint a group of entities in $KG_2$ that may correspond to a specific entity in $KG_1$. The candidate generation function can be depicted as follows:

 $$G(e_1, KG_1, KG_2) = M \; \; \forall \, e_1 \in Sig(KG_1)$$


Next, the filtering function $F$ evaluates the quality of the generated candidate matches. It achieves this by leveraging the mapping confidence and employing cardinality rules or thresholds to establish a criterion for accepting or rejecting a candidate mapping. Furthermore, filtering functions may also "repair" alignments by mitigating logical inconsistencies between mappings and signatures, thereby enhancing their coherence. In essence, these functions eliminate mappings that exhibit low quality, violate cardinality constraints, or introduce incoherence. We can represent this process symbolically as follows:

$$F(M, KG_1, KG_2) = K^M $$


The complete mapping function $f$ is derived by combining the candidate generation function(s) $G$ with the filtering function(s) $F$, leading to a final set of high-quality candidate matches. Consequently, the global correspondence function can be expressed as:
 $$f(KG_1, KG_2) = F(G(e_1, KG_1, KG_2)) \;\; \forall \, e_1 \in Sig(KG_1)$$

 In this setting, the objective is to maximize the number of mappings while preserving the coherence of $KG_M$. Given that maximizing the number of mappings and maintaining coherence are two contradicting strategies, they can either be optimized under an assumption of correctness or uncertainty.
 
An assumption of correctness implies that flawless logic governs the set of axioms constituting $KG_1$ and $KG_2$. Consequently, any incoherent mapping cannot be accurate. Under this assumption, the optimization problem can be formulated as follows:
 $$\arg\max_{f} \sum_{e_1\in Sig(KG_1)} \; \sum_{f(e_1) \subset Sig(KG_2) \, \vee \, |f(e_1)| \leq t} consist(e_1) * consist(f(e_1)) * c(e_1, f(e_1)) $$

where $f(e_1)$ is the set of entities mapped to $e_1$ under $f$, $c$ is a function that returns the confidence value of a given mapping, and $t$ is a cardinality threshold.  Therefore, the goal is to maximize the mapping confidence for a given cardinality of coherent mappings.


On the other hand, under an assumption of uncertainty, axioms might be incorrect or incomplete. By viewing logical incoherence as a result of incorrect or incomplete information, we can conceive of a heuristic that aims to minimize logical incoherence while maximizing the confidence value. This heuristic should, in practice, deal with the trade-off between relying on explicit structural information (axioms) and relying on all other sources of information (\eg lexical/semantic, internal/external). The simplest form of this heuristic, denoted as $\mathcal{H}$, would involve a threshold $\theta$ beyond which the confidence value surpasses logical incoherence. Under this setting, the optimization problem can be formulated as such:

 $$\arg\max_{f} \sum_{e_1\in Sig(KG_1)} \; \sum_{f(e_1) \subset Sig(KG_2) \, \vee \, |f(e_1)| \leq t} \begin{cases}c(e_1, f(e_1)) & c(e_1, f(e_1)) \geq \theta\\consist(e_1, f(e_1)) * c(e_1, f(e_1)) & c(e_1, f(e_1)) < \theta\end{cases} $$


 An alternative perspective on $\mathcal{H}$ involves adopting a "soft" approach towards addressing unsatisfiability. In this regard, the previously defined $softconsist$ function can be employed to formulate the optimization problem as follows:

 $$\arg\max_{f} \sum_{e_1\in Sig(KG_1)} \; \sum_{f(e_1) \subset Sig(KG_2) \, \vee \, |f(e_1)| \leq t} softconsist(e_1) * softconsist(f(e_1)) * c(e_1, f(e_1)) $$

Within this setup, a mapping's confidence will be subject to a penalty if its constituent entities are involved in mappings that are unsatisfiable, and the penalty can be proportional to the number of unsatisfiability cases it is related to.

\subsubsection{Supervised Setting}\label{sup_learning}

Supervised learning approaches can be applied to KG alignment when a partial alignment is available, typically manually produced and deemed correct (\ie a reference alignment). In this setting, the solution may still be divided into a generator and a filter, particularly when combining supervised and unsupervised algorithms. For instance, we could use an unsupervised generator with a supervised filter. However, this division is not strictly necessary. It is entirely possible to attempt to learn the problem holistically.

Let us assume there is a reference alignment symbolized by a function $ref$, which calculates the truth value for a given relationship between two entities, such that $ref(e_1, e_2, r) \in \{0, 1\}$ and where $r \in \{\sqsubseteq, \sqsupseteq, \equiv\}$ is a semantic relation. In this setting, the correspondence function $f$ can be optimized problem to maximize $ref$. This optimization problem can be expressed as follows:

 $$\arg\max_{f} \sum_{e_1\in Sig(KG_1)} \; \sum_{f(e_1) \subset Sig(KG_2) \, \vee \, |f(e_1)| \leq t} ref(e_1, f(e_1))$$


The goal is to find the correspondence function $f$ that maximizes the sum of similarities between the corresponding entities in the two KGs.




\subsection{Overview/State of the Art}\label{SOA_align}


We follow Euzenat \& Shvaiko's \cite{euzenat2013} classification of ontology matching techniques into eight specific categories according to the type of input they take. Content-based matching techniques place their focus on the internal information obtained from the ontologies being matched, while context-based techniques focus on external information.

\emph{Context-based approaches} comprise two categories defined by their use of either formal or informal resources.

\textbf{Formal resource-based approaches} use formal resources such as previous alignments and other ontologies to either discover new candidate mappings or filter out unlikely candidates. A notable illustration of this approach is put forth by Scharffe \etal \cite{scharffe2014}. Traditional matching systems, such as AML \cite{faria2013}, and several others likewise harness reference alignments and ontological resources within their frameworks. This category also encompasses the majority of supervised or semi-supervised methods, such as VeeAlign \cite{iyer2021}, as they draw upon reference alignments for training purposes.



\textbf{Informal resource-based approaches} also rely on external resources, but instead of formal resources like ontologies, they utilize informal resources such as annotated relationships between entities or text. A prevalent method involves employing a thesaurus such as thesaurus.com or subject-specific equivalents. For instance, LogMap \cite{jimenez2011} leverages the UMLS lexicon to expand synonyms. This category also encompasses language models like BERTMap~\cite{he2022}, which undergo pre-training on large amounts of external textual content.


\emph{Content-based approaches} can be categorized into six distinct groups based on their usage of KG content. These methodologies encompass leveraging linguistics across the KG, capitalizing on the structural attributes of the KG, directing attention towards specific instances, and harnessing the semantic richness embedded within the KGs.


\textbf{String-based approaches} involve gauging string similarity within entity names and descriptions to generate plausible candidate mappings. These methods, often referred to as lexical matchers, are prominently featured in classical matching systems like AML \cite{faria2013}, and LogMap \cite{jimenez2011}. A variety of string distance metrics, encompassing Jaccard, Euclidean, Levenshtein, Jaro-Winkler, Jensen-Shannon, and others \cite{cohen2003}, can be employed by these matchers. String-based techniques are primarily geared towards identifying candidate mappings rather than refining them, as they inherently carry a level of uncertainty. This lack of certainty renders them better suited for assessing the likelihood of a mapping. To discover potential mappings, these techniques can be applied both within the ontology itself and across external resources.


\textbf{Language-based approaches}, unlike string-based approaches, do not treat names or descriptions as mere strings. Instead, they consider them as partial words or sentences in a language by leveraging natural language processing techniques. Prominent examples of these approaches are those that rely on large language models pre-trained on extensive corpora, such as BERTMap \cite{he2022}. 


\textbf{Constraint-based approaches} compare the constraints applied to the definition of entities, including constraints on cardinality, types, attributes, or keys. The underlying assumption is that equivalent entities will exhibit similar constraints. Typically, these techniques are used in conjunction with others, as similar constraints on entities are often a common characteristic. Constraint-based approaches are commonly employed as filters to exclude mappings that violate specific constraints, such as cardinality. Examples of such filters can be found in most matching systems \cite{megdiche2016}.


\textbf{Graph-based approaches} leverage the inherent graph structure of KGs and ontologies to generate mappings. State-of-the-art approaches in this category often rely on Graph Neural Networks (GNNs). Examples of such approaches include GCN-align \cite{wang2018}, MEDTO \cite{Hao2021}, and AliNet \cite{Sun2020}.

\textbf{Instance-based approaches} assess the similarity between classes based on the instances associated with them. The underlying intuition is that similar classes will have similar instances. Several data analysis techniques can be employed to compare instances as proposed by Isaac \etal \cite{isaac2007}. However, a major challenge with these approaches is the existence of sparse instances for comparison in some cases.


\textbf{Model-based approaches} leverage the semantic interpretation of paired ontologies by employing formal reasoning techniques, such as description logics reasoning. Due to their precise, albeit computationally intensive, nature, these techniques are primarily used as "repair" algorithms. Their main objective is to filter out logically incoherent mappings from a pre-selected set of candidate mappings. Several works have been published in this area, such as: Solimando \etal \cite{solimando2017}, Santos \etal \cite{santos2015}, Jimenez \etal \cite{jimenez2012, jimenez2013}, Shvaiko \etal \cite{shvaiko2009}, and Meilicke \etal \cite{meilicke2007}.


Until now, we have discussed individual matching techniques. However, state-of-the-art ontology alignment systems typically consist of multiple matching techniques working together. Evaluating the performance of these systems is commonly done through the Ontology Alignment Evaluation Initiative (OAEI) \cite{OAEIWebsite, pour2023}. OAEI is an annual competition that includes various tracks designed to evaluate different tasks and components related to ontology alignment. 


While many systems participate in the OAEI tracks, two systems have consistently demonstrated excellent performance over the years: AML \cite{faria2013} and its successor, Matcha \cite{faria2022}, as well as LogMap \cite{jimenez2011}. Despite the competition from other systems, AML/Matcha and LogMap have maintained their superiority across most tracks for more than a decade. As a result, these systems have become the go-to choices for ontology alignment, and any neurosymbolic approaches aiming to make an impact in this field will need to surpass their performance.

Of particular significance for the neurosymbolic field is the new Bio-ML track \cite{he2022b}. This track is the first to emphasize machine learning-based systems for ontology alignment, making it a crucial arena for neurosymbolic approaches. In the inaugural edition of this track, Matcha-DL \cite{faria2022}, a combination of Matcha (a rule-based system) with a simple neural network and BERTMap \cite{he2022} achieved the best results. These two systems currently serve as the baseline for neurosymbolic approaches to strive for and surpass in performance. 


\section{Neuro-Symbolic Methods}\label{NSai}


Neurosymbolic AI, also known as Neuro-symbolic computing, is a subfield of artificial intelligence (AI) that envisions integrating symbolic and subsymbolic approaches to this field. While these approaches differ significantly at their extremes (\eg expert systems vs. multi-layer perceptron), advancements in both domains are slowly bridging the longstanding gap. Although the AI field has, in recent years, predominantly emphasized subsymbolic approaches, such as deep learning techniques, there is a growing consensus that augmenting subsymbolic AI with certain properties of classical symbolic systems could address its current limitations \cite{garcez2023, hitzler2022, kautz2022}.

Symbolic approaches are those that leverage high-level or explicit knowledge representations, such as those expressed through formal logic and other formal languages, along with symbol manipulation, to accomplish their objectives. During the early stages of AI, classical forms of these approaches gained significant popularity \cite{feigenbaum1963}. By employing a formal knowledge source, possess the capability to generate new knowledge and validate assumptions through logical deduction precisely. Since they operate based on manipulations of symbols that are understandable to humans, their deductions can be easily verified even by non-experts. However, symbolic approaches face challenges due to their reliance on human experts to encode knowledge into formal rules, especially when these experts need to represent complex and predominantly continuous domains as discrete rules. This task is often exceedingly difficult, if not impossible. In fact, as early as 1950, Turing expressed his scepticism about the feasibility of representing the world in such a manner in his seminal paper 'Computing Machinery and Intelligence' \cite{turing2009}. Due to these limitations, pure symbolic approaches such as the CYC project \cite{Douglas1995} have largely fallen out of favour, giving rise to the prominence of subsymbolic approaches.

Subsymbolic systems did not garner significant attention from the artificial intelligence community until the emergence of deep learning and big data. This lack of attention primarily stemmed from limitations in computing power and the availability of adequate data to effectively explore these approaches. It was not until the emergence of crucial datasets like WordNet \cite{miller1995} and ImageNet \cite{deng2009} that more robust exploration became feasible. In contrast to symbolic techniques, deep learning focuses on learning through induction from informal data sources, known as datasets. Deep learning models approximate unknown target functions by learning from numerous instances of representations of entities. They can then resort to these approximated functions to make predictions on unseen instances that fall within the learned distribution. When the training dataset is sufficiently representative of the domain being explored, these techniques can achieve exceptional performance \cite{lecun2015}.

The advantage of deep learning models lies in their ability to automatically extract crucial features from raw data, eliminating the need to construct a comprehensive model of the world to solve a given problem. They have achieved remarkable results in various domains such as natural language processing \cite{ vaswani2017, brown2020}, computer vision \cite{ren2015, kipf2016, He2016}, game playing \cite{silver2016, silver2017}, and multimodal learning \cite{Tadas2019}. This naturally led to a shift towards subsymbolic approaches that has gained significant traction in recent years, acknowledging the need for more flexible and adaptable AI systems capable of handling complex and unstructured information. Similar to the perception surrounding symbolic systems, where it was believed that any problem could be solved with enough rules, some initially speculated that logical reasoning and consciousness would emerge with sufficient data and computing power.

However, as deep learning models approach their limits in terms of scale, they face challenges in simple commonsense reasoning tasks that can be effortlessly solved by infants. Questions are now being raised about whether this approach alone provides the complete answer \cite{garcez2023}. Deep learning models also encounter issues with data inefficiency, generalisation beyond the training distribution, fairness concerns, and, perhaps most notably, a lack of human comprehensibility \cite{hitzler2022}.

The integration of subsymbolic and symbolic approaches initially encountered challenges due to experts in both fields who disregarded the potential of each other's techniques. Some individuals continue to advocate exclusively for pure subsymbolic approaches, dismissing any symbolic involvement. Fortunately, within the field of AI, an increasing number of researchers now recognize the potential of integrating these approaches to address their respective deficiencies \cite{garcez2023, hitzler2022, kautz2022}.

One lens through which the differences between symbolic and subsymbolic techniques can be viewed, particularly in a neuroscience context, is the framework of System I and System II proposed by Kahneman in his book "Thinking, Fast and Slow" \cite{Kahneman2000}. Kahneman suggests that the human brain employs two types of thinking: a fast, unconscious, automatic, and effortless thinking attributed to System I, which responds to stimuli, and a slow, logical, and effortful thinking attributed to System II, which is utilized to solve complex problems. Notably, this theory aligns with the distinctions between subsymbolic and symbolic techniques, indicating that achieving human-like intelligence may involve integrating these methodologies. Artificial neural networks can be viewed as abstractions of the inner workings of the human brain, while formal logic systems serve as abstractions of the perceptual and reasoning processes of our conscious mind \cite{garcez2023}. 

By recognizing the value of both subsymbolic and symbolic approaches and understanding the potential benefits of integration, the field of Neurosymbolic AI is paving the way for advancing artificial intelligence toward more comprehensive and human-like intelligence.

\subsection{Integrating Symbolic and Subsymbolic Approaches}


On the one hand, deep learning systems demonstrate exceptional performance in training on raw data, showcasing robustness against errors and outliers. However, they encounter challenges when it comes to incorporating expert knowledge, comprehending logical reasoning, and achieving interpretability. On the other hand, symbolic systems are easily interpretable and capable of leveraging expert knowledge, but they are susceptible to errors and outliers in the knowledge source and pose difficulties in the training process. This naturally raises the question of how we can effectively integrate these two abstractions. In \cite{hitzler2022} Hitzler \etal present intriguing ideas and insights on approaching this integration.

One suggested proposition is the use of pure subsymbolic models to solve classical symbolic problems, such as algebraic problems, logical deduction and abduction, rule learning, term rewriting, or planning. These challenging problems, even for complex symbolic systems, provide an opportunity to test the limits of current deep learning approaches. 

Another avenue proposed is the incorporation of symbolic knowledge bases, such as KGs and other forms of explicit metadata, to enhance subsymbolic systems. Neurosymbolic models could leverage KGs in combination with deep learning models to improve zero-shot learning, as shown in the work by Lee \etal \cite{lee2018}. Other approaches, such as K-Bert \cite{liu2020} and the model proposed by Ma \etal \cite{ma2019}, fuse KGs with large language models to provide domain context, thereby enhancing coherence and consistency within a given domain.

A further suggested path involves the use of background knowledge, such as KGs, to explain the input-output behaviour of black-box neural networks. Numerous works have delved into this area, exploiting KGs or other forms of expert knowledge. Examples of such research include the works by Sarker \etal \cite{sarker2017}, Geng \etal \cite{geng2019, geng2021}, and Dalal \etal \cite{dalal2023}.

Lastly, a key research avenue in the field of Neurosymbolic AI involves the coupling of deep learning models with symbolic components. This approach aligns closely with the objectives of Neurosymbolic AI, aiming to integrate subsymbolic and symbolic models. To provide a framework for understanding the different levels of integration between symbolic and subsymbolic methodologies, Kautz \cite{kautz2022} proposed a taxonomy for neurosymbolic systems. This taxonomy is further explored by Garcez and Lamb \cite{garcez2023}.


Within this taxonomy, Type I integration, referred to as \textbf{Symbolic Neurosymbolic} integration, represents the use of standard deep learning techniques. The argument for considering this as neurosymbolic lies in the fact that deep learning techniques often begin with symbols (\eg natural language processing), encode them into a continuous space, process this vector space using a neural engine, and then decode the result back into symbols, thereby demonstrating Symbolic Neurosymbolic integration.

A Type II integration, known as \textbf{Symbolic[Neuro]} integration, involves loosely coupled hybrid systems that use a core neural model within a symbolic problem solver. This type of integration is exemplified by systems like DeepMind's AlphaGo \cite{silver2016}. 

Type III neurosymbolic integration, referred to as \textbf{Neuro|Symbolic} integration, entails a neural model and a symbolic reasoning system working together to achieve a complex goal by handling complementary tasks. In these hybrid systems, a neural model receives non-symbolic input and produces a symbolic data structure, which is then processed by a symbolic reasoner to verify complex statements. The Neuro Symbolic Concept Learner \cite{mao2019} serves as a prime example of this type of integration. 

A Type IV neural-symbolic system (\textbf{Neuro:Symbolic $\rightarrow$ Neuro} integration) involves a standard neural network with symbolic constraints on training, achieved either by training on symbolic datasets or by implementing constraints on the learning process itself. Arabshahi \etal \cite{arabshahi2018} propose an approach to learning and reasoning over mathematical programs, which is then extended in \cite{arabshahi2019} to an architecture capable of extrapolating to more challenging symbolic reasoning problems. Garcez and Lamb \cite{garcez2023} also argue that systems with learning constraints, such as those explored in Logical Neural Networks (LNN) \cite{riegel2020}, should fall under this type of integration. LNNs establish a one-to-one correspondence between neurons and logical formulas, where each neuron's activation function can be constrained to approximate a logical operation. The main challenges associated with these networks lie in mapping one-to-one networks for large-scale problems and integrating them with standard deep learning architectures.

Type V systems (\textbf{Neuro\_\{Symbolic\}} integration) are the most tightly coupled without achieving full integration. In these systems, symbolic rules are mapped onto embeddings that serve as regularizers for the network's loss functions. Current examples of this integration include tensorization methods like LTNs \cite{badreddine2022} and Tensor Product Representations \cite{smolensky2016}. LTNs combine deep neural networks with first-order fuzzy logic by embedding elements of a formal language in a vector space. Unlike LNNs, LTNs enable full integration with other neural architectures for solving complex problems. LTNs are trained on facts and rules and can infer novel logical statements from data. Recent versions of LTNs have demonstrated excellent performance across various AI tasks, such as multi-label classification, relational learning, data clustering, semi-supervised learning, regression, embedding learning, and query answering \cite{badreddine2022}. Wagner and Garcez \cite{wagner2021} also show that LTNs can significantly improve fairness metrics while maintaining state-of-the-art performance by incorporating fairness constraints into the neural engine.

Lastly, Type VI integration, known as \textbf{Neuro[Symbolic]} integration, refers to fully integrated systems where a symbolic engine is embedded into a neural network, representing a comprehensive integration of System 1 and System 2. Kautz suggests that such architectures should allow for combinatorial reasoning, potentially by leveraging attention-based approaches. While fully integrated systems do not currently exist, Lamb \etal \cite{lamb2020} argue that integrating graph neural networks with Neurosymbolic AI and incorporating attention mechanisms may be a promising direction towards achieving this type of system.

\section{Challenges and Opportunities}



\begin{table}
\caption{Table matching KGA challenges to Neuro-Symbolic approaches that offer promising potential solutions.} \label{tab:challenges}
\begin{tabular}{@{}ccc@{}}
\toprule
\textbf{Challenges}     & \textbf{Mapping}                                 & \textbf{Repair}                                                                                                                                     \\ \toprule
Tackling Large Size     & Large-Language Models (BERTMap)           & Integration of mapping and repair (Adversarial Training)                                                                                        \\ \midrule
Handling Heterogeneity  & Formal knowledge injection (K-BERT, Instruction Training)              & \begin{tabular}[c]{@{}c@{}}Subsymbolic approximators\\ (MLP, GCN)\end{tabular}                                                                      \\ \midrule
Achieving Correctness     & Formal knowledge injection (K-BERT, Instruction Training)                                                & \begin{tabular}[c]{@{}c@{}}Soft Repair (LNN, LTN)\\ Integration of mapping and repair (Adversarial Training)\end{tabular}                       \\ \midrule
Minimizing Human Effort & Data efficient approaches (K-Bert, Instruction Training, LTN)   & \begin{tabular}[c]{@{}c@{}}Human-in-the-loop approaches  (RL)\end{tabular}          \\ \midrule
Explaining Decisions    & \multicolumn{2}{c}{\begin{tabular}[c]{@{}c@{}}Post-hoc explanations (concept induction, path-based explanations)\\ Tightly coupled NSai (LNN, LTN)\end{tabular}}   \\ \bottomrule
\end{tabular}
\end{table}

Table \ref{tab:challenges} outlines the correlation between existing challenges within KGA and emerging Neuro-Symbolic AI approaches. These approaches are classified based on their primary functions: mapping or repair.



\subsection{Tackling Large Size}

One of the current challenges facing the KGA field is scalability. As KGs continue to grow in size, symbolic systems that often rely on complex sets of formal rules and require meticulous fine-tuning became increasingly harder to scale to larger problems. Furthermore, when applied to different KGs, these systems essentially start from scratch, resulting in impractically long run times for very large KGs and rendering them unsuitable for real-time applications.

Prominent examples like AML \cite{faria2013} and LogMap \cite{jimenez2011} highlight the challenges faced by pure symbolic systems.  While they continue to achieve state-of-the-art performance in specific domains and have shown effectiveness over the years, they lack some of the advantages offered by more data-driven approaches.


KG alignment systems that explore \textbf{Large Language Models} (LLM), like BERTMap \cite{he2022}, undergo initial training on extensive corpus and subsequent fine-tuning using reference alignments for various KGs. Although the initial training phase may require considerable time, the subsequent inference process benefits from the efficiency of deep learning approaches, allowing them to outperform symbolic approaches in terms of speed when matching new, unseen KGs. This natural scalability can lead to improved performance on larger and more complex KGs while maintaining the ability to rapidly infer mappings. Therefore, LLMs offer advantages for scaling KGA system's mapping process, as the inference time experiences only negligible changes as the KGs grow in size and complexity.

Further efficiency can be achieved by tackling a common time-consuming feature of contemporary KGA systems: the separation of mapping and repair processes. Given the fundamental requirement of most repair algorithms, outlined in section \ref{problem_def}, for a complete alignment, state-of-the-art KGA systems divide the task into distinct mapping and repair phases. This approach introduces the possibility of redundant efforts between mapping and repair algorithms.

To tackle this issue, we propose an approach for the  \textbf{integration of mapping and repair algorithms within an adversarial training framework}, such as a simple generator-discriminator architecture. This architecture would allow the learning and execution of mapping and repair functions in tandem, streamlining the KGA process to a single inference task across KGs. While optimal performance might still require two inference runs $-$ one for source-to-target and another for target-to-source $-$ this approach holds promise for more efficient scaling on larger KGs compared to existing state-of-the-art systems.

A similar architecture was already introduced by \cite{Qu2019}, showing promising results. Notably, this approach employs relatively straightforward generator and discriminator networks in contrast to the complex mapping and repair algorithms used by competing systems. Exploring whether this approach's efficiency endures when applied to more complex KGA tasks would be an interesting avenue for further research.

\subsection{Handling Heterogeneity}

Addressing heterogeneity across diverse domains is of crucial importance in the context of KGA challenges. Real-world KGs exhibit considerable disparity stemming from a multitude of origins, domains, and even languages. Notably, the most intriguing KGA tasks involve the alignment of exceptionally dissimilar KGs. In order to effectively manage such diversity, a substantial degree of generalisation ability becomes imperative.



Subsymbolic algorithms can effectively leverage patterns inherent in raw data to produce high-quality mappings, while symbolic algorithms heavily rely on expert-defined rules for generating mappings. This has led to the emergence of LLM-based methods, such as BERTMap \cite{he2022}, which are capable of effectively generalising across domains. Their scale enables them to store and leverage a significant amount of information within their networks. This, in theory, should mean that deep learning models have the potential to continuously enhance their generalisation power as more data becomes available.




However, the very characteristics that contribute to the success of LLMs can also be limitations. These models heavily rely on large corpora and numerous reference mappings to perform well, creating requirements that are not always feasible to meet. Additionally, training these models effectively demands substantial computational power. Much of the knowledge embedded in the structure of KGs, which could potentially be inferred, must instead be induced from thousands of examples, resulting in an overall data-inefficient model.


Neurosymbolic integration offers a promising pathway for achieving efficient data usage. \textbf{Formal knowledge injection} models like K-Bert \cite{liu2020} demonstrate comparable data induction capabilities to pure symbolic models while incorporating symbolic knowledge bases, resulting in improved performance when faced with the out-of-distribution challenges associated with smaller datasets. This integration reduces the reliance on raw data, making it a valuable approach for KGA problems.

Another emerging avenue for leveraging symbolic knowledge within LLM frameworks involves zero-shot learning via instruction-based training. This innovative approach has gained prominence more recently, facilitated by the subsequent wave of LLMs that followed in the footsteps of ChatGPT. It entails the refinement of general language models like GPT-3 \cite{brown2020} for specific tasks, achieved through instructions and illustrative examples provided as textual context. This technique circumvents direct alterations to the model's parameters through supervised or semi-supervised learning, thereby enabling the democratization of LLM-based research.

In the context of KGA tasks, this novel fine-tuning methodology could be effectively applied by incorporating the KG's structural information into the mapping process, such as an entity's subclasses. OntoLAMA \cite{he2023} explores a similar approach for subsumption tasks, yielding promising results. While the adaptation of such a technique to complex matching and equivalence tasks might not follow a straightforward trajectory, we hold the belief that it presents a promising avenue for research, warranting further exploration.

In the domain of repair, acquiring the skill to correct these mappings could be facilitated by the use of \textbf{subsymbolic approximators}, trained on incomplete reference mappings. Notably, this approach has had encouraging results, as showcased by the MatchaDL system \cite{faria2013}. In this system, a neural network functions as a heuristic, learning to adjust the weighting of scores generated by symbolic algorithms. Further advancements in this domain could explore the potential of GNNs for learning these weights, considering the structural positions of entities within their respective graphs.

\subsection{Achieving Correctness}

Another major challenge of KGA tasks is the requirement for models that can effectively reason over and interpret KGs to achieve correct alignments. Meaning that the mappings between entities not only need to be semantically accurate but also logically satisfiable when reasoning over KGs.

KG alignment symbolic systems commonly employ separate algorithms for scoring mappings, followed by a heuristic to consolidate these scores and generate candidates. 
While these techniques perform well individually, they lack integration, as information sharing among algorithms is limited. The final heuristic used for consolidation becomes crucial while often relying on domain experts' rules, which may not be optimal and can vary in performance across different domains.

On the other hand, subsymbolic systems employ complex networks that can combine multiple architectures to explore several properties of the KG when generating candidate mappings. GNNs can capture structural information similarly to symbolic algorithms, while LLMs, as discussed earlier, can learn from large corpora and make better semantic interpretations. Recent subsymbolic works that explore this approach to mapping, such as \cite{he2022, Qu2019, chen2021}, have achieved state-of-the-art performance in certain tasks. However, subsymbolic models have limitations, such as their difficulty in leveraging logical reasoning and semantics in commonsense reasoning tasks. As such, without incorporating formal forms of external knowledge, their potential to fully replace symbolic approaches in mapping tasks is limited. 


Neurosymbolic integration can play a crucial role in harnessing the advantages of both symbolic and subsymbolic systems. As we have shown previously, approaches like instruction training and context incorporation into the attention mechanism of language models (as exemplified by K-Bert \cite{liu2020}) can be effectively employed to assimilate structured knowledge from aligned KGs or external references into the mapping procedure. Moreover, we have also demonstrated that subsymbolic approximators exhibit the capacity to rectify discrepancies in a range of symbolic and subsymbolic matching algorithms.


Nevertheless, this type of loosely coupled neurosymbolic approach might not be enough to achieve correctness. Real-world KGs often contain errors, including incomplete information or incompatibilities among KGs.

Let's consider the following example \cite{jimenez2011b}, from the FMA and NCI ontologies, henceforward denoted as $O_{FMA}$ and $O_{NCI}$. In $O_{FMA}$, "Lymphokine" is expressed as $Lymphokine \sqsubseteq Protein$, while in $O_{NCI}$, an homologous "Therapeutic\_Lymphokine"  is expressed as $Therapeutic\_Lymphokine \sqsubseteq Pharmacologic\_Substance$. Additionally, a disjointedness axiom $\alpha: Protein \sqcap Pharmacologic\_Substance \sqsubseteq \bot$ is defined in $O_{NCI}$. Current algorithms would typically generate two candidate mappings: $<Protein \in O_{FMA} \equiv Protein \in O_{NCI}>$ and $<Lymphokine \in O_{FMA} \equiv Therapeutic\_Lymphokine \in O_{NCI}>$. However, state-of-the-art logical reasoners (repair algorithms) would discard one of these mappings due to the logical inconsistency derived from $\alpha$. In reality, both mappings are correct, and the problem lies with either $\alpha$, or the compatibility between the ontologies.

Tightly coupled neurosymbolic systems, such as LNNs \cite{riegel2020} or LTNs \cite{badreddine2022}, could offer a promising approach to handle logical satisfiability more flexibly. These systems integrate adjustable logical constraints into their learning procedure, allowing the learning of an acceptable level of logical unsatisfiability. Under the idea of soft and hard constraints in the repair process explored in section \ref{problem_def}, these approaches could present a promising heuristic to learn soft constraints in the repair process, effectively producing a \textbf{soft repair} 

Currently, the practical applications of LNNs \cite{riegel2020} are limited due to constraints in current implementations. However, they present intriguing research avenues for KGA by mapping KG information to the network architecture and incorporating adjustable constraints on logical behaviour. While LNNs heavily rely on receiving formal knowledge as input, which may pose challenges in other applications, it aligns well with KGA, where the inputs are predominantly symbolic data structures. Nevertheless, scalability issues related to network mapping remain a concern for the field. Overcoming implementation challenges should be prioritized in future research into these models, as it can pave the way for more effective and efficient neuro-symbolic models.

The current version of LTNs \cite{badreddine2022} appears more feasible for immediate application in KGA problems. Due to their tighter integration with subsymbolic models, they can be more easily combined with state-of-the-art subsymbolic systems. A fundamental research direction is to evaluate their performance in large-scale settings and assess their ability to reduce the amount of raw data and computational power required to achieve state-of-the-art results.

An interesting research direction for tightly coupled neurosymbolic systems would be to explore combining the advantages of clearly defining adjustable logical constraints, such as in LNNs, with the practical integration found in tensorization-based methods like LTNs.


An alternative avenue to address the challenge of ensuring correctness involves revisiting the integration between mapping and repair functions. The current dilemma of achieving correctness stems from the segmentation of mapping generation and logical filtering tasks. This division, while commonly practised, carries inherent limitations as it assumes the correctness of KGs. This assumption can undermine the quality of alignment due to its overly restrictive nature. Therefore the adoption of an adversarial training framework is, yet again, a potentially promising approach. This idea becomes particularly compelling when considering the use of a soft repair heuristic, such as an LNN \cite{riegel2020} or an LTN \cite{badreddine2022}, as the discriminator. The interplay between mapping and repair networks under this framework could potentially yield significant advancements in addressing the problem of correctness in alignment~tasks.


\subsection{Minimizing Human Effort}

The challenge of minimizing human effort within the domain of KGA is intricately linked to the notion of streamlining systems to require as little human intervention as possible. Present-day state-of-the-art systems still rely on human involvement during both the mapping and repair phases.

In the context of the mapping process, human intervention primarily centres on dataset curation and manual calibration. This rings especially true for symbolic systems, where formal rules require meticulous curation by domain experts and heuristics demand manual tuning. As a result, these systems demand significant human effort to operate and maintain, rendering them challenging for non-experts to use. Despite this, contemporary machine learning systems have also failed to provide significant improvement in terms of human effort. This paper stresses that these systems often suffer from severe sample inefficiency, demanding copious examples even for elementary tasks. While pre-training has mitigated the need for extensive curated samples in large models such as LLMs, the inefficiency of fine-tuning remains notable, especially considering the large amount of effort required to generate reference alignments for model training.

As discussed earlier, \textbf{data efficient approaches} emphasizing sample efficiency and tapping into formal knowledge within subsymbolic models, like zero-shot learning through instruction training or approaches like that proposed in K-Bert \cite{liu2020}, present potential avenues for alleviating human effort in the mapping phase. These approaches can potentially reduce the size of the reference alignments and external curated data needed to achieve state-of-the-art results.

Shifting the focus to the repair phase, human intervention traditionally occurs toward the end of the alignment process. Here, certain systems flag mappings with low confidence for validation by human experts. While sometimes indispensable, this verification process introduces hurdles when striving to seamlessly integrate these algorithms into plug-and-play applications. In response, we advocate for the exploration of alternative interactive methodologies. One intriguing avenue is \textbf{human-in-the-loop reinforcement learning (HLRL)} \cite{hejna2023, christiano2017, lee2021}. HLRL leverages human feedback to refine reward functions, demonstrating success, as seen in the case of ChatGPT's chat functionalities, which were trained using a similar approach based on the GPT-3 base generator model \cite{brown2020}. Within the domain of KGA, this feedback-driven function could be integrated into an inclusive KGA system, either as part of a discriminator in an adversarial training setup or as a reward in a reinforcement learning framework. Regardless of the specific integration strategy, learning a human preference function empowers the automation of expert feedback, eventually leading to the realization of a fully automated system.

\subsection{Explaining Decisions}


Explainability is another significant challenge in KGA tasks, as automated algorithms often produce non-transparent results, making it difficult for end-users to understand the alignment decisions or identify any inconsistencies on biases in the results. This limits end-users' trust in the results of KGA.

From the outset of deep learning, it became clear that explaining the results of these black box models would be a challenge. Practical applications of these algorithms for decision-making require some comprehension of the reasoning behind the models, but in most cases, understanding the decision-making of deep learning models is a challenging task even for experts. This task becomes almost impossible when we consider large language models with billions of activation parameters. In the last few years, some approaches have risen to prominence to analyze the activation patterns of neural networks \cite{ribeiro2016, Lundberg2017}, but this task remains very complex, and human interpretable explanations are only attainable in some select cases. 

Although symbolic systems inherently offer more interpretability compared to subsymbolic approaches, they can still be challenging for non-experts to grasp if decision explanations are not readily generated.

This challenge has not gone unnoticed by the semantic web community, which has proposed strategies to tackle this challenge \cite{pesquita2021}. A particularly intriguing approach, especially for explaining deep learning models, involves \textbf{Post-Hoc Explanations}. This methodology leverages external formal knowledge to shed light on the decisions made by black box models. Within post-hoc explanations, several approaches were already explored. Particularly pertinent to KGA are Concept Induction \cite{widmer2022, dalal2023} and Path-based explanations \cite{xiong2017, Zhu2022}. 

Concept Induction employs a symbolic reasoner to attach formal knowledge to neural activations. This approach facilitates comprehension of subsymbolic constituents that constitute a deep learning model. This aligns closely with the KGA domain, where formal knowledge is already abundant and can be seamlessly augmented.

Path-based explanations focus instead on generating logical pathways or "reasoning lines" across ontologies to formulate plausible rationales for a given decision. This proves especially valuable for KGA since ontologies already exist for reasoning purposes, and this knowledge foundation can be augmented, if necessary, with supplementary formal resources. These approaches offer the added advantage of being model agnostic, meaning they explain the decision outcome rather than interpreting the model itself. This flexibility enables their use for both symbolic and subsymbolic systems.





\textbf{Tightly coupled neurosymbolic algorithms} offer another promising solution to the limitations of purely neural or symbolic models. By incorporating the reasoning power of symbolic models, they become inherently capable of generating explanations for their decisions. The most notable examples of tightly coupled neurosymbolic networks with this inherent explainability are LNNs \cite{riegel2020} and LTNs \cite{badreddine2022}. These networks can offer explanations irrespective of their size or complexity, paving a promising path for the development of explainable systems.

\section{Outlook}


The wake of deep learning has rekindled interest in neurosymbolic integration. This paper presents a comprehensive overview of how this promising approach can enhance the performance of state-of-the-art KGA algorithms. By bridging the gap between symbolic reasoning and data-driven learning, neurosymbolic integration offers a promising pathway to overcome many of the current challenges in KGA.

We have outlined several critical challenges in the KGA and promising research paths for unlocking them by leveraging neurosymbolic methods. These paths comprise a spectrum of innovative strategies, such as: using symbolic knowledge within subsymbolic systems; incorporating soft logical constraints into the repair process; exploring the potential of tightly coupled hybrid systems; and integrating the mapping and repair process into a single framework. Each of these directions represents an opportunity to improve the performance of KGA algorithms in one or more of the key challenges identified.

While our paper showcases several approaches, each focusing on specific aspects of KGA tasks, we envision future research trajectories in this domain to be even more ambitious. Beyond whether individual approaches can successfully tackle specific challenges, we advocate for research endeavours that transcend singular methodologies. The aspiration is to witness the emergence of KGA systems that combine several neurosymbolic approaches, culminating in a system that collectively surpasses prevailing state-of-the-art algorithms. This consolidation of methodologies holds the promise of reshaping the landscape of KGA research and significantly propelling the field forward.


\bibliographystyle{ios1}           
\bibliography{bibliography}        

\end{document}